\documentclass[a4paper,11pt]{article}

\usepackage{booktabs}

\usepackage{lineno}
\usepackage[colorlinks=false]{hyperref}
\usepackage{bm}
\usepackage{graphicx}
\usepackage{amssymb}
\usepackage{amsmath}
\usepackage{tabularx}
\usepackage{natbib}
\usepackage{floatrow}
\usepackage{caption}
\usepackage{multirow}
\usepackage{amsmath}
\usepackage{amsfonts}
\usepackage{amssymb}
\usepackage{xcolor}
\usepackage{subfig}
\usepackage{graphicx,amsmath,bm}
\usepackage[binary-units=true]{siunitx}
\usepackage{array}
\usepackage{authblk}
\usepackage{rotating}
\usepackage{enumerate}
\usepackage{ragged2e}

\usepackage[left=15mm,right=15mm,top=1.5cm,bottom=1.5cm,includeheadfoot]{geometry}
\setcitestyle{square,numbers}
\begin{document}
	
	\title{Structured Multi-Level-Set Inversion for 3-D Electromagnetic Inverse Scattering}
	
	\author[1]{Yutong~Du}
	\author[1]{Zicheng~Liu}
	\author[1]{Bo~Qi}
	\author[1]{Yali~Zong}
		\author[1]{Xin~Li}
	\author[1]{Peixian~Han}
	
	\affil[1]{\scriptsize Department of Electronic Engineering, School of Electronics and Information, Northwestern Polytechnical University, Xi'an 710029, China}
	\maketitle
	
	\abstract{
		Many dielectric targets encountered in 3-D electromagnetic inverse scattering exhibit piecewise-homogeneous material distributions, which provide useful structural information for inversion. To exploit this property, this paper proposes a structured multi-level-set inversion (SMLSI) method within a physics-driven neural network (PDNN) framework. Instead of directly parameterizing the contrast distribution, SMLSI represents it in terms of object support and material distribution. Multiple level-sets are mapped to soft-support functions and combined through a soft-union operation to construct the object support. The normalized contributions of these soft-support functions, together with trainable complex relative permittivities, determine the material distribution within the support. The level-set residuals are parameterized by a coordinate-based neural network, and the trainable parameters are optimized from scattered field data. Structure-aware regularization terms are introduced to promote binary object support and spatially coherent material regions while reducing material mixing and redundant activation of level-set supports. Adaptive regularization weighting is further employed to reduce the dependence on manual parameter tuning. Numerical experiments demonstrate that SMLSI improves reconstruction accuracy over direct contrast parameterization and enables the recovery of irregular 3-D geometries, closely spaced objects, and multiple material regions. Compared with CSI and $L_{2/3}$-FBE-WCIE, SMLSI achieves more accurate reconstructions on representative synthetic targets. Experimental validation using 3-D Fresnel data further demonstrates its applicability to measured scattering data.}
	
	\section{Introduction}
	Einverse scattering problems (ISPs) \cite{chen2018computationalEMIS} aim to reconstruct the spatial distribution of dielectric properties from measured scattered fields. The problems have attracted sustained attention in microwave imaging \cite{Ahmed2021MicroSec}, nondestructive testing \cite{Mostafa2026NDT,li20213}, biomedical diagnosis \cite{Yan2025Medi,Qin2022medical}, geophysical exploration \cite{Esposito2024ExpVal}, and radar remote sensing \cite{Liu2026SAR}. However, electromagnetic inverse scattering is inherently nonlinear and ill-posed, making the stable recovery of material properties from scattered fields challenging, particularly for targets with irregular geometries, sharp interfaces, and composite structures. At the same time, many dielectric targets can be characterized by piecewise-homogeneous material distributions, which provide useful structural prior information for inversion.
	
	Classic iterative methods, such as the distorted Born iterative method (DBIM) \cite{chew1990DBIM,haddadin1995DBIM}, contrast source inversion (CSI) \cite{peter1997CSI,richard2001CSI}, and subspace optimization methods (SOM) \cite{chen2010SOM,chen_2010SOM,pan2011SOM}, have been widely investigated for electromagnetic inverse scattering. Various regularization techniques, such as Tikhonov regularization, total variation (TV), edge-preserving penalties, and sparsity-promoting $\ell_p$ quasi-norm regularization, have also been incorporated into these frameworks to stabilize the inversion \cite{Xu2018HybridRegular,Liu2023HybridRegular,Li2026FBE3D}. These methods generally represent the unknown dielectric contrast by spatially distributed values. 
	
	Learning-based approaches provide another route to solving ISPs. Supervised data-driven neural networks learn mappings from network inputs (e.g., scattered fields or an initial estimate) to contrast distributions and can provide rapid inference after offline training \cite{wei2018BPS,Li2018DeepNIS}. Physics-enhanced data-driven neural networks further incorporate electromagnetic information into network inputs, architectures, or training objectives \cite{wei2019PhaNN,Liu2022PhaGuiNN,Liu2022SOMnet,Tao2023NNBIM,Du2025QuaDNN}. More recently, physics-driven neural network (PDNN) solvers use neural networks to parameterize unknown physical quantities and optimize the network parameters directly from measured scattered field data~\cite{Song2022uSOM,Du2025PDNN,Du2026IPDNN,Du2026PDF}. These solvers reduce or eliminate the dependence on labeled training data and offer flexible parameterizations of the unknown quantities.
	
	Level-set methods provide a natural representation for piecewise-homogeneous targets by describing material interfaces through the zero level sets of continuous level-set functions~\cite{Dorn2006LS,Dorn2000LS}. Colgan et al.~\cite{Colgan2015LS} used a single level-set function to distinguish two predefined tissue classes. Shah and Moghaddam~\cite{Shah2018LS} developed a fast level-set method and extended it to multimaterial microwave imaging, while Shah et al.~\cite{Shah20193DLS} further developed a 3-D level-set method for joint shape and dielectric contrast recovery. These studies demonstrate the effectiveness of level-set formulations for incorporating piecewise-homogeneous structure into inverse scattering. However, previous level-set formulations mainly use level sets to describe geometric boundaries or predefined material partitions, with the material properties either prescribed or estimated separately from the shape update. Building on these studies, this work proposes a structured multi-level-set inversion (SMLSI) method that replaces direct contrast parameterization with a structured representation of object support and material distribution.
	
	The main contributions are summarized as follows.
	
	1) A structured multi-level-set representation is proposed for 3-D inverse scattering. Multiple level-set functions are combined to construct the overall object support, while their normalized contributions and trainable complex relative permittivity parameters determine the material distribution within the support. The number of level-set functions serves as a representation capacity and does not need to match the number of objects or material regions.
	
	2) Structure-aware regularization terms are developed for the proposed representation to promote binary object support and spatially coherent material regions while suppressing material mixing and redundant activation of level-set functions. An adaptive weighting strategy is further introduced to balance these regularization terms during inversion and reduce the dependence on manually selected regularization strengths.
	
	3) A staged physics-driven optimization strategy is developed to progressively optimize the geometric and material parameters of the structured representation from the measured scattered fields.
	
	The remainder of this paper is organized as follows. Section~\ref{sec:formulateISPs} formulates the considered 3-D electromagnetic inverse scattering problem. Section~\ref{sec:InvScheme} presents the formulation and implementation of the proposed SMLSI. Section~\ref{sec:NumSet} analyzes the structured representation through sensitivity and ablation studies. Section~\ref{sec:Performance} evaluates the reconstruction performance through synthetic experiments, comparisons with existing methods, and experimental validation. Finally, Section~\ref{sec:Conclu} concludes the paper.
	
	\section{Formulation of Inverse Scattering Problems}
	\label{sec:formulateISPs}
	\begin{figure}
		\centering
		\includegraphics[width = .3\linewidth]{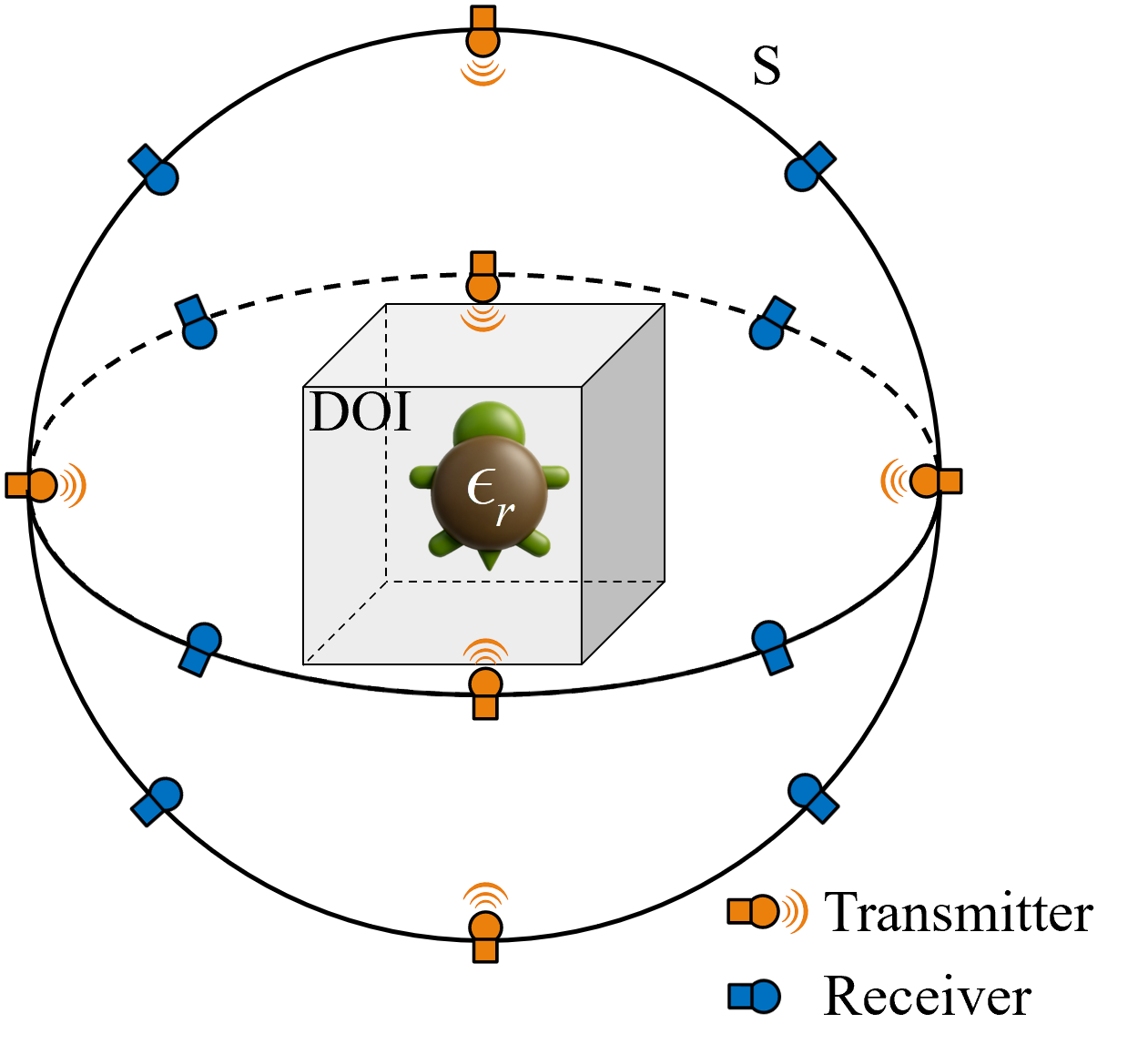}
		\caption{Schematic of the considered 3-D imaging configuration.}
		\label{fig:ISPs}
	\end{figure}
	The considered 3-D imaging configuration is illustrated in Figure~\ref{fig:ISPs}. The domain of interest (DOI) is sequentially illuminated by incident fields generated by $N_i$ transmitters, and the scattered fields are measured by $N_s$ receivers located on the observation surface $S$ surrounding the DOI.
	
	The total electric field is governed by the vector volume integral equation \cite{Schaubert1984VIE,Roger1961THEMF}, which can be expressed as
	\begin{equation}
		\mathbf E^{\mathrm{tot}}(\mathbf r) = \mathbf E^{\mathrm{inc}}(\mathbf r) + k_0^2 \int_\text{DOI} {\mathbf G}(\mathbf r,\mathbf r') \mathbf J(\mathbf r') d\mathbf r', \quad \mathbf r\in \text{DOI} ,
		\label{eq:state_equation_3-D}
	\end{equation}
	where $\mathbf r$ denotes the observation position in the DOI. $\mathbf E^{\mathrm{inc}}$ and $\mathbf E^{\mathrm{tot}}$ denote incident and total electric fields, respectively. $k_0$ is the background wavenumber, and ${\mathbf G}(\mathbf r,\mathbf r')$ denotes the dyadic Green's function \cite{tai1994dyadic}. The induced current is defined as
	\begin{equation}
		\mathbf J(\mathbf r) = \boldsymbol\chi(\mathbf r)\mathbf E^{\mathrm{tot}}(\mathbf r),
		\label{eq:contrast_source}
	\end{equation}
	where $\boldsymbol\chi(\mathbf r)=\boldsymbol\epsilon_r(\mathbf r)-1$ is the contrast function, $\boldsymbol\epsilon_r(\mathbf r)$ being the relative permittivity.
	
	The scattered electric field on the observation surface is radiated by the induced current source and is expressed as
	\begin{equation}
		\mathbf E^{\mathrm{sca}}(\mathbf r) = k_0^2\int_\text{DOI}{\mathbf G}(\mathbf r,\mathbf r')\mathbf J(\mathbf r')d\mathbf r',
		\quad \mathbf r\in \text{S}.
		\label{eq:data_equation_3-D_vector}
	\end{equation}
	
	The inverse scattering problem aims to reconstruct the unknown contrast distribution $\boldsymbol{\chi}$ from the measured $\mathbf E^{\mathrm{sca}}$ by solving the following optimization problem
	\begin{equation}
		\min_{\boldsymbol{\Theta}}
		\mathcal L(\boldsymbol{\Theta})
		=
		\mathcal L_{\mathrm{data}}
		\big(\hat{\boldsymbol{\chi}}(\boldsymbol{\Theta})\big)
		+
		\mathcal R(\boldsymbol{\Theta}).
		\label{eq:general_inverse_problem}
	\end{equation}
	$\mathcal L_{\mathrm{data}}$ is the term constraining the data discrepancy, $\mathcal R$ is the regularization term imposing prior constraints on the desired solution, and $\boldsymbol{\Theta}$ denotes the set of trainable model and representation parameters.	
	
	\begin{figure*}
		\centering
		\includegraphics[width = \linewidth]{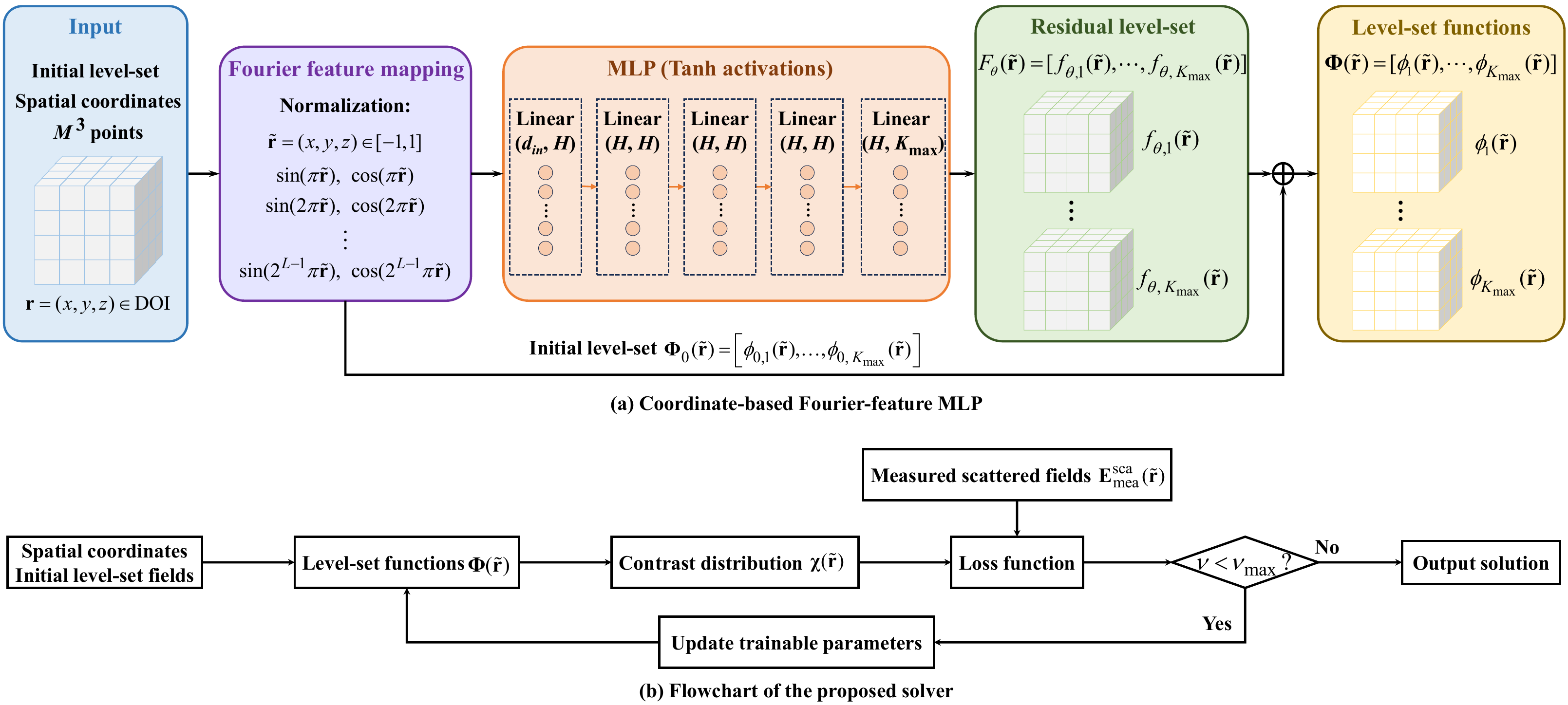}
		\caption{Sketch of the proposed SMLSI. (a) Architecture of the Coordinate-based Fourier-feature MLP. (b) Flowchart of the iteration scheme.}
		\label{fig:Overview}
	\end{figure*}
	
	\section{Structured Multi-Level-Set Inversion}
	\label{sec:InvScheme}
	
	The overall framework of the proposed structured multi-level-set inversion (SMLSI) is illustrated in Figure~\ref{fig:Overview}. Instead of directly parameterizing the contrast distribution, SMLSI constructs it from a set of level-set functions. As shown in Figure~\ref{fig:Overview}(a), a coordinate-based Fourier-feature multi-layer perceptron (MLP) predicts the residuals of the level-set functions, which are combined with the initial level-sets to obtain the current level-set functions. The corresponding contrast distribution is then constructed from these level-set functions. As shown in Figure~\ref{fig:Overview}(b), the predicted scattered fields are computed from the current contrast distribution and used to evaluate the loss function, after which the trainable parameters are updated iteratively. The level-set functions and contrast distribution are recomputed after each update, and the procedure continues until the iteration index $\nu$ reaches the maximum number of iterations $\nu_{\max}$. The detailed level-set parameterization, contrast representation, objective function, and optimization strategy are presented in the following subsections.
	
	\subsection{Multi-Level-Set Parameterization}
	\label{subsec:MLSP}
	
	The DOI is discretized into an $M\times M\times M$ inversion grid, yielding $M^3$ spatial points. Let $\tilde{\mathbf r}\in[-1,1]^3$ denote the normalized coordinate corresponding to the spatial coordinate $\mathbf r$, obtained by
	\begin{equation}
		\tilde r_d=2\frac{r_d-r_d^{\min}}{r_d^{\max}-r_d^{\min}}-1,\qquad d\in\{x,y,z\}.
	\end{equation}
	To represent the geometry of the unknown target, $K_{\max}$ level-set functions are introduced as
	\begin{equation}
		\phi_k(\tilde{\mathbf r})=\phi_{0,k}(\tilde{\mathbf r})
		+f_{\theta,k}(\tilde{\mathbf r}),\qquad
		k=1,\ldots,K_{\max},
		\label{eq:phi_residual}
	\end{equation}
	where $\phi_{0,k}$ is the initial level-set function and $f_{\theta,k}$ is a learnable residual function. The zero level set $\phi_k(\tilde{\mathbf r})=0$ defines the represented boundary, $\phi_k(\tilde{\mathbf r})<0$ denotes the region inside the $k$th level set, whereas $\phi_k(\tilde{\mathbf r})>0$ the region outside it. Since a single level-set function can represent multiple disconnected regions, $K_{\max}$ is not necessarily equal to the number of objects to be reconstructed.
	
	The initial level-set function is chosen as a spherical signed-distance function,
	\begin{equation}
		\phi_{0,k}(\tilde{\mathbf r})=\|\tilde{\mathbf r}-\mathbf c_k\|_2-R_0,
		\label{eq:initial_phi}
	\end{equation}
	where $\mathbf c_k$ denotes the initial center of the $k$th level-set function and $R_0$ is the initial radius. The numerical setting of $K_{\max}$ and the initialization of the corresponding $\mathbf c_k$ and $R_0$ are discussed in Section~\ref{subsec:paraSets}.
	
	The residual functions $f_{\theta,k}$ are parameterized by a coordinate-based multi-layer perceptron (MLP). The normalized coordinates are embedded by a Fourier feature mapping,
	\begin{equation}
		\gamma(\tilde{\mathbf r})=[\tilde{\mathbf r},\sin(\pi\tilde{\mathbf r}),\cos(\pi\tilde{\mathbf r}),\ldots,\sin(2^{L-1}\pi\tilde{\mathbf r}),
		\cos(2^{L-1}\pi\tilde{\mathbf r})],
		\label{eq:fourier_feature}
	\end{equation}
	where the sine and cosine functions are computed element-wise and $L$ denotes the number of frequency bands. The resulting Fourier features are then fed into the MLP, whose outputs are
	\begin{equation}
		F_{\theta}(\tilde{\mathbf r})
		=\operatorname{MLP}_{\theta}\!\left(\gamma(\tilde{\mathbf r})\right)
		=[f_{\theta,1}(\tilde{\mathbf r}),f_{\theta,2}(\tilde{\mathbf r}),\ldots,f_{\theta,K_{\max}}(\tilde{\mathbf r})].
		\label{eq:mlp_output}
	\end{equation}
	Each output $f_{\theta,k}$ provides the residual of the \(k\)th level-set function at position \(\tilde{\mathbf r}\). The level-set functions over the DOI are then obtained from \eqref{eq:phi_residual}.
	
	\subsection{Structured Contrast Representation}
	\label{subsec:CR}
	
	The level-set functions obtained in Section~\ref{subsec:MLSP} are first mapped to soft-support functions. This mapping converts the unbounded level-set values into bounded variables in $[0,1]$, providing an approximation to the binary support indicator. Specifically, the $k$th level-set function is mapped as
	\begin{equation}
		q_k(\tilde{\mathbf r}) = \sigma\!\left[-\beta\phi_k(\tilde{\mathbf r})\right],
		\label{eq:qk}
	\end{equation}
	where $\sigma(x)=1/[1+\exp(-x)]$ denotes the sigmoid function and $\beta>0$ controls the steepness of the sigmoid mapping w.r.t. the level-set value. Accordingly, $q_k$ approaches $1$ inside the $k$th level-set region, approaches $0$ outside it, and equals $0.5$ on the zero level set $\phi_k(\tilde{\mathbf r})=0$.
	
	\begin{figure}
		\centering
		\includegraphics[width = .38\linewidth]{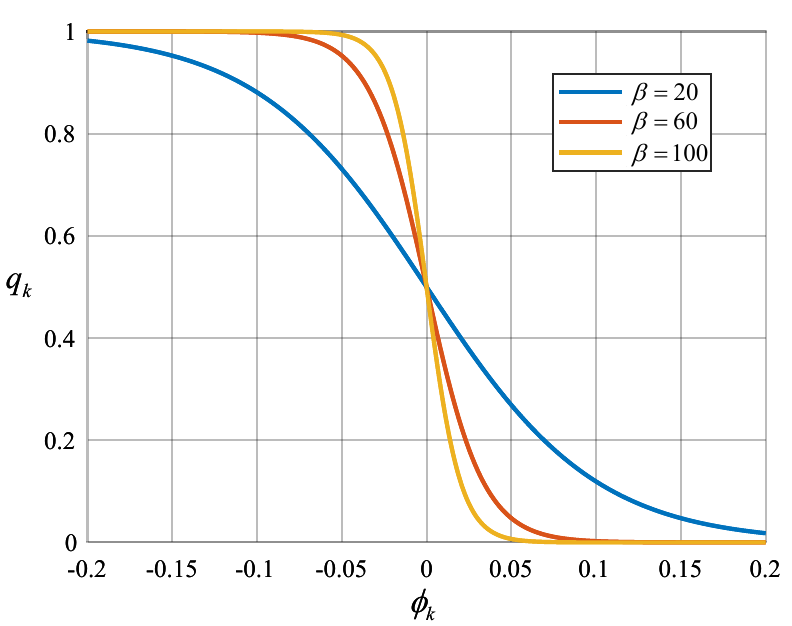}
		\caption{Mapping from the level-set function $\phi_k$ to the soft-support function $q_k$ for several illustrative values of $\beta$. A small $\beta$ gives a smoother transition, whereas a large $\beta$ yields a sharper transition closer to binary support.}
		\label{fig:phikqk}
	\end{figure}
	
	A small $\beta$ yields a smooth transition and stable gradients, whereas a larger $\beta$ produces a closer approximation to the binary support. During optimization, $\beta$ is updated at iteration $\nu$ according to
	\begin{equation}
		\beta^{(\nu)}=\min\left\{80,\,5\times1.2^{\left\lfloor \nu/50\right\rfloor}\right\}.
	\end{equation}
	As illustrated in Figure~\ref{fig:phikqk}, the gradual increase in $\beta$ narrows the transition region and makes $q_k$ progressively closer to a binary indicator function.
	
	To adjust the contribution of each soft-support function $q_k$, a learnable scalar parameter $\eta_k$ is introduced, and
	\begin{equation}
		\tilde q_k(\tilde{\mathbf r})=\sigma(\eta_k) q_k(\tilde{\mathbf r}),
		\label{eq:gated_q}
	\end{equation}
	where $\sigma(\eta_k)\in(0,1)$ acts as a learnable gate controlling the contribution of the $k$th level-set function.
	
	The global object support is then constructed by combining the weighted soft supports $\tilde q_k$. A direct summation of multiple supports may produce values larger than one in overlapping regions, leading to amplification of the support value. To preserve a bounded soft-support representation, a soft-union operation is therefore adopted,
	\begin{equation}
		q_{\mathrm{union}}(\tilde{\mathbf r})
		=1-\prod_{k=1}^{K_{\max}}
		\left[1-\tilde q_k(\tilde{\mathbf r})\right].
		\label{eq:q_union}
	\end{equation}
	Since $0\leq \tilde q_k\leq1$, the resulting support always satisfies
	$0\leq q_{\mathrm{union}}\leq1$. When one or more $\tilde q_k$ take large values at a given location, $q_{\mathrm{union}}$ approaches $1$, indicating a high likelihood that the location belongs to the reconstructed object.
	
	To determine the material distribution within the global object support, the weighted soft supports are normalized as
	\begin{equation}
		p_k(\tilde{\mathbf r})=
		\frac{\tilde q_k(\tilde{\mathbf r})}
		{\displaystyle\sum_{l=1}^{K_{\max}}\tilde q_l(\tilde{\mathbf r})+\delta},
		\label{eq:pk}
	\end{equation}
	where $\delta=1\times10^{-8}$ is introduced for numerical stability. The normalized weight $p_k$ quantifies the contribution of the $k$th trainable complex permittivity parameter
	$\boldsymbol\epsilon_{r,k}$ at position $\tilde{\mathbf r}$. The reconstructed complex contrast is given by
	\begin{equation}
		\chi(\tilde{\mathbf r})=q_{\mathrm{union}}(\tilde{\mathbf r})\sum_{k=1}^{K_{\max}}p_k(\tilde{\mathbf r})\left(\boldsymbol\epsilon_{r,k}-1\right).
		\label{eq:chi_mls}
	\end{equation}
	Thus, $q_{\mathrm{union}}$ determines the overall object geometry, whereas $p_k$ and $\boldsymbol\epsilon_{r,k}$ determine the spatial material properties within the reconstructed support. Moreover, a single level-set function can represent multiple disconnected regions, while several level-set functions may jointly contribute to the material property at the same spatial location.
	
	\subsection{Objective Function}
	\label{subsec:SR}
	The reconstruction objective function consists of a data-fidelity term and four structure-aware regularization terms designed for the structured multi-level-set representation. The objective function is written as
	\begin{equation}
		\mathcal{L}=\mathcal{L}_{\mathrm{Data}}+\lambda_{\mathrm{TV}}\mathcal{R}_{\mathrm{TV}}+\lambda_{\mathrm{Bin}}\mathcal{R}_{\mathrm{Bin}}+\lambda_{\mathrm{Gini}}\mathcal{R}_{\mathrm{Gini}}+\lambda_{\mathrm{Act}}\mathcal{R}_{\mathrm{Act}},
		\label{eq:loss}
	\end{equation}
	where $\lambda_{\mathrm{TV}}$, $\lambda_{\mathrm{Bin}}$, $\lambda_{\mathrm{Gini}}$ and $\lambda_{\mathrm{Act}}$ are the corresponding regularization weights.
	
	The data-fidelity term quantifies the normalized mismatch between the measured and predicted scattered fields,
	\begin{equation}
		\mathcal L_{\mathrm{Data}}=\frac{ \|{\mathbf{E}}^{\mathrm{sca}}_{\mathrm{mea}}-\hat{\mathbf{E}}^{\mathrm{sca}}(\hat{\boldsymbol{\chi}})\|_2^2}{\|{\mathbf{E}}^{\mathrm{sca}}_{\mathrm{mea}}\|_2^2
		}.
		\label{eq:loss_data}
	\end{equation}
	Here, $\hat{\mathbf E}^{\mathrm{sca}}(\hat{\boldsymbol{\chi}})$ denotes the scattered field predicted from the reconstructed contrast distribution $\hat{\boldsymbol{\chi}}$.
	
	The regularization terms are defined as
	\begin{subequations}
		\label{eq:loss_terms}
		\begin{align}
			&\mathcal{R}_{\mathrm{TV}}=\frac{1}{K_{\max}}\sum_{k=1}^{K_{\max}}\mathrm{TV}(q_{\mathrm{union}}(\tilde{\mathbf{r}})p_k(\tilde{\mathbf{r}})),\\
			&\mathcal{R}_{\mathrm{Bin}} = \frac{1}{M^3}\sum_{i=1}^{M^3}q_{\mathrm{union}}(\tilde{\mathbf{r}}_i)[1-q_{\mathrm{union}}(\tilde{\mathbf{r}}_i)],\\
			&\mathcal{R}_{\mathrm{Gini}} = \frac{1}{M^3}\sum_{i=1}^{M^3}q_{\mathrm{union}}(\tilde{\mathbf{r}}_i)\sum_{k=1}^{K_{\max}}p_k(\tilde{\mathbf{r}}_i)[1-p_k(\tilde{\mathbf{r}}_i)],\\
			&\mathcal{R}_{\mathrm{Act}} = \frac{1}{M^3}\sum_{i=1}^{M^3}\sum_{k=1}^{K_{\max}}\tilde{q}_k(\tilde{\mathbf{r}}_i).
			\label{eq:loss_act}
		\end{align}
	\end{subequations}
	Here, the total variation operator is defined as
	$\mathrm{TV}(u)=\langle|\nabla_x u|\rangle+\langle|\nabla_y u|\rangle+\langle|\nabla_z u|\rangle$. $\nabla_x$, $\nabla_y$, and $\nabla_z$ denote the first-order finite-difference operators along the three coordinate directions, and $\langle\cdot\rangle$ denotes the arithmetic average over all entries. The term $\mathcal{R}_{\mathrm{TV}}$ is applied to $q_{\mathrm{union}}p_k$ instead of only $q_{\mathrm{union}}$, thereby promoting the spatial continuity of individual material regions in multi-material reconstruction. The binarization term $\mathcal{R}_{\mathrm{Bin}}$ penalizes nonbinary support values, thereby encouraging $q_{\mathrm{union}}$ to approach a binary object-support function. The Gini-type impurity term \(\mathcal{R}_{\mathrm{Gini}}\), analogous to the Gini impurity used in classification and regression trees~\cite{breiman1984classification}, encourages the normalized weights $p_k$ to approach either zero or one, thereby reducing the mixing of different candidate permittivities at each spatial location within the reconstructed support. The activation term $\mathcal{R}_{\mathrm{Act}}$ penalizes the total spatial occupancy of the level-set supports, thereby discouraging unnecessary or spatially redundant activation. These regularization terms therefore provide complementary constraints on spatial continuity, support binarization, material assignment, and component activation. Their effectiveness is further evaluated through the ablation study in Section~\ref{subsec:Ablation}.
	
	\subsection{Optimization Strategy}
	\label{subsec:OS}
	The inversion is divided into three stages by two iteration thresholds, $\nu_w$ and $\nu_p$, with $\nu_w<\nu_p$. For $\nu<\nu_w$, only the level-set parameters $\theta$ and the trainable parameters $\{\eta_k\}_{k=1}^{K_{\max}}$ controlling the contribution of each level-set function are updated using the data-fidelity term, while the complex relative permittivity parameters $\{\epsilon_{r,k}\}_{k=1}^{K_{\max}}$ remain fixed. At $\nu=\nu_w$, the regularization terms begin to contribute to the optimization, and their weights are optimized together with $\theta$ and $\{\eta_k\}_{k=1}^{K_{\max}}$. The permittivity parameters remain fixed during this stage. From $\nu=\nu_p$ onward, the permittivity parameters are updated jointly with the other trainable parameters.
	
	For convenience, the regularization terms are denoted as
	\begin{equation}
		\{\mathcal R_i\}_{i=1}^{4}=\{\mathcal R_{\mathrm{TV}}, \mathcal R_{\mathrm{Bin}}, \mathcal R_{\mathrm{Gini}}, \mathcal R_{\mathrm{Act}}\},
		\label{eq:Ri}
	\end{equation}
	with the corresponding regularization weights $\{\lambda_i\}_{i=1}^{4}=\{\lambda_{\mathrm{TV}}, \lambda_{\mathrm{Bin}}, \lambda_{\mathrm{Gini}}, \lambda_{\mathrm{Act}}\}$.
	
	At $\nu=\nu_w$, the regularization weights are initialized according to the current magnitudes of the data-fidelity and regularization terms
	\begin{equation}
		\lambda_i^{(0)}=\min\left\{\frac{\rho\mathcal L_{\mathrm{Data}}^{(\nu_w)}}
		{4\left(\mathcal R_i^{(\nu_w)}+\delta\right)}, \lambda_{\max}\right\}, \qquad i=1,2,\ldots,4,
		\label{eq:lambda_init}
	\end{equation}
	where $\delta=1\times10^{-8}$ is introduced for numerical stability, $\lambda_{\max}$ is the prescribed upper bound, and $\rho$ controls the initial contribution of the regularization terms relative to the data-fidelity term.
	
	After initialization, the regularization weights are allowed to adapt during the inversion. To ensure positivity, each weight is parameterized as $\lambda_i=\exp(\alpha_i)$. Directly treating $\lambda_i$ as a trainable parameter in $\lambda_i\mathcal R_i$ would lead to a degenerate solution in which the objective is reduced by driving $\lambda_i$ toward zero. To avoid this problem, an auxiliary term is introduced
	\begin{equation}
		\Phi_{\lambda}=-\sum_{i=1}^{4}c_i\log\lambda_i,
		\label{eq:Phi_lambda}
	\end{equation}
	where $c_i=\lambda_i^{(0)}\mathcal R_i^{(\nu_w)}$, and \(c_i\) remains fixed throughout the subsequent optimization. The objective function is  therefore written as
	
	\begin{equation}
		\mathcal L^{(\nu)}=
		\begin{cases}
			\mathcal L_{\mathrm{Data}}^{(\nu)},
			& \nu<\nu_w,\\[1mm]
			\mathcal L_{\mathrm{Data}}^{(\nu)}
			+\displaystyle\sum_{i=1}^{4}
			\lambda_i\mathcal R_i^{(\nu)}
			+\Phi_{\lambda},
			& \nu\geq\nu_w.
		\end{cases}
		\label{eq:stage_loss}
	\end{equation}
	
	For $\nu\geq\nu_w$, the terms in \eqref{eq:stage_loss} that depend on the \(i\)th regularization weight are $\lambda_i\mathcal R_i-c_i\log\lambda_i$. Since \(\alpha_i=\log\lambda_i\), differentiating these terms w.r.t. $\alpha_i$ gives $\lambda_i\mathcal R_i-c_i$. When the upper-bound constraint is inactive, stationarity w.r.t. $\alpha_i$ therefore requires $\lambda_i\mathcal R_i=c_i$. At $\nu=\nu_w$, since $c_i=\lambda_i^{(0)}\mathcal R_i^{(\nu_w)}$, this condition is satisfied by $\lambda_i=\lambda_i^{(0)}$. During the subsequent inversion, the adaptive optimization tends to compensate for changes in the magnitude of $\mathcal R_i$ through $\lambda_i$.
	
	To prevent excessively large regularization weights when the
	corresponding regularization terms become small, the constraint
	$0<\lambda_i\leq\lambda_{\max}$ is imposed. Since
	$\lambda_i=\exp(\alpha_i)$, the upper bound is enforced after each
	gradient update by
	\begin{equation}
		\alpha_i
		\leftarrow
		\min\{\alpha_i,\log\lambda_{\max}\}.
		\label{eq:lambda_projection}
	\end{equation}
	
	The trainable parameters are updated according to the optimization schedule using the gradients of \eqref{eq:stage_loss}. For $\nu\geq\nu_w$, the projection in \eqref{eq:lambda_projection} is applied after each gradient update of $\alpha_i$. The inversion continues until the maximum number of iterations $\nu_{\max}$ is reached. The numerical settings used in the synthetic experiments are given in Section~\ref{subsec:paraSets}.
	
	\section{Numerical Setup and Analysis of the Structured Representation}
	\label{sec:NumSet}
	
	\begin{figure}
		\centering
		\includegraphics[width = .52\linewidth]{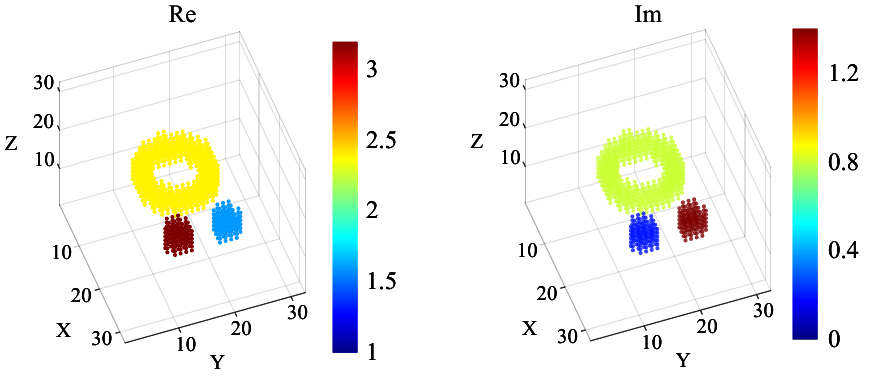}
		\caption{Ground truth of the multi-material Austria target.}
		\label{fig:TruthMulti}
	\end{figure}
	
	\begin{figure*}
		\centering
		\includegraphics[width = \linewidth]{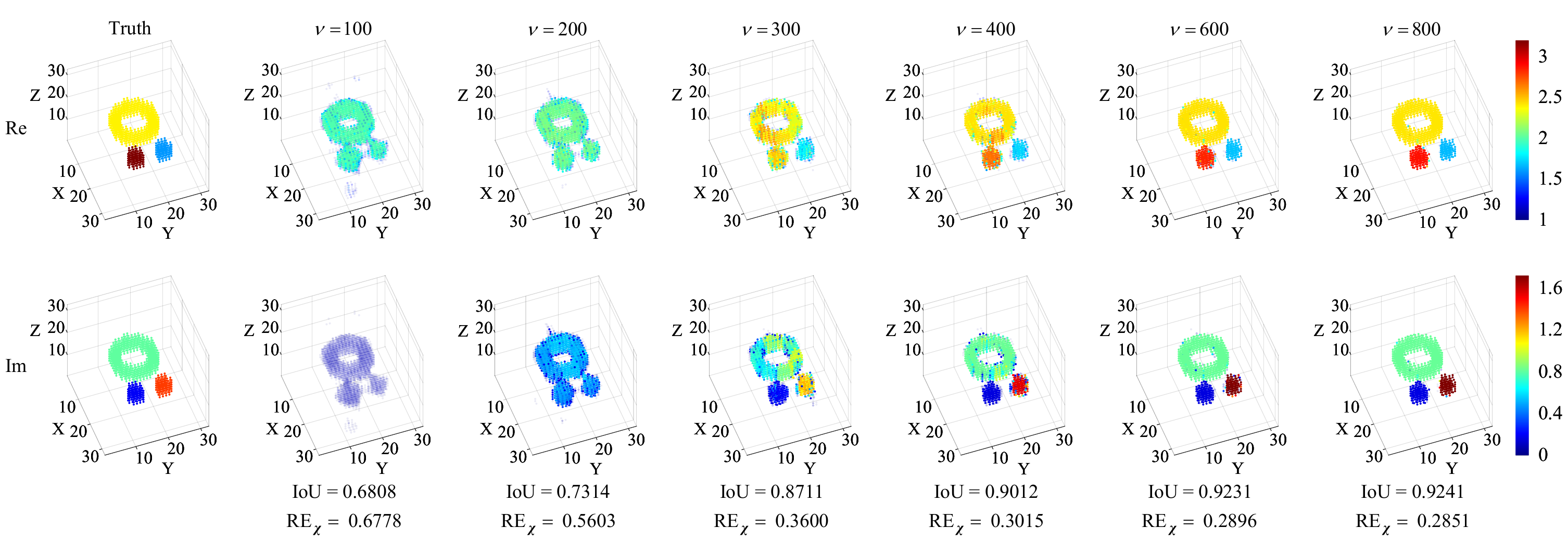}
		\caption{Evolution of the reconstructed relative permittivity distribution during the inversion for the multi-material Austria target. The corresponding IoU and $\mathrm{RE}_{\chi}$ values are given below each iteration.}
		\label{fig:qunion}
	\end{figure*}
	
	\subsection{Numerical Setup and Evaluation Metrics}
	\label{subsec:paraSets}
	
	Unless otherwise specified, all synthetic experiments in Sections~\ref{sec:NumSet} and~\ref{sec:Performance} use the following numerical configuration. The DOI is a cube with a side length of $0.15 \mathrm{m}$. The reconstruction is performed on a $32\times32\times32$ inversion grid, i.e., $M=32$. The synthetic scattered field data are generated using the method of moments (MoM)~\cite{Ney1985MoM,LAKHTAKIA1992MoM}. The operating frequency is 4 GHz. A total of $N_i=16$ transmitters and $N_s=32$ receivers are uniformly distributed on a spherical surface centered at the DOI with a radius of $20\lambda$, where $\lambda$ denotes the wavelength in free space. A single co-polarized measurement channel is considered, where the incident electric fields are polarized along the local azimuthal unit vector $\hat{\boldsymbol{\phi}}=(-\sin\phi,\cos\phi,0)$, and the scattered electric fields are projected onto the corresponding local azimuthal direction at each receiver. Therefore, each inversion uses a $32\times16$ complex-valued scattered field matrix. Noise-free data are used by default, while additive white Gaussian noise is added to the scattered fields in the noise-robustness experiments at the specified SNR levels.
	
	For the structured representation, the initial level-set centers are generated from a 3-D Sobol sequence~\cite{Sobol1967} to obtain a deterministic and spatially well-distributed initialization without requiring prior knowledge of the target location. The initial radius and the number of level-set functions are set to $R_0=0.12$ and $K_{\max}=4$, respectively, and the sensitivity to $K_{\max}$ is examined in Section~\ref{subsec:SenstoCI}.
	
	The residual level-set functions are represented by a coordinate-based MLP with four hidden layers, each containing 64 neurons and using tanh activation functions. The Fourier feature mapping in \eqref{eq:fourier_feature} uses $L=4$ frequency bands. The learnable gates are initialized as $\sigma(\eta_k)=0.8$, and all complex relative permittivity parameters are initialized as $\varepsilon_{r,k}=2.0+0.1\mathrm{j}$.
	
	The optimization strategy described in Section~\ref{subsec:OS} is adopted in all numerical experiments, with $\nu_w=100$ and $\nu_p=150$. The Adam optimizer is used for all trainable parameters. The initial learning rates are set to $1\times10^{-2}$ for the level-set parameters $\theta$ and the gating parameters $\{\eta_k\}_{k=1}^{K_{\max}}$. For the complex relative permittivities $\{\boldsymbol\epsilon_{r,k}\}_{k=1}^{K_{\max}}$, the initial learning rate is set to $1\times10^{-1}$. The parameters in \eqref{eq:lambda_init} are set to $\lambda_{\max}=1$ and $\rho=0.05$. The learning rates are halved every 150 iterations. The maximum number of iterations is set to $\nu_{\max}=800$. All inversion methods are implemented and evaluated on a workstation equipped with 128 GB RAM, a 3.2 GHz i9 CPU and an NVIDIA GeForce RTX 4090 GPU.
	
	All analyses in Section~\ref{sec:NumSet} are performed using the same 3-D multi-material Austria target shown in Figure~\ref{fig:TruthMulti}. The target consists of an annular region and two smaller regions with different complex relative permittivities.
	
	To evaluate the overall reconstruction accuracy, the relative error (RE) of the reconstructed complex contrast is defined as
	\begin{equation}
		\mathrm{RE}_{\boldsymbol{\chi}}=\frac{\left\|\hat{\boldsymbol{\chi}}-\boldsymbol{\chi}\right\|_2}{\left\|\boldsymbol{\chi}\right\|_2},
		\label{eq:RE_chi}
	\end{equation}
	where $\hat{\boldsymbol{\chi}}$ and $\boldsymbol{\chi}$ denote the reconstructed and ground truth contrast distributions, respectively.
	
	Since the proposed representation separates object geometry from material properties, the reconstruction of the object support is additionally evaluated using the intersection over union (IoU). Let $\mathcal{D}$ denote the ground truth support, and obtain the reconstructed binary support from the soft support $q_{\mathrm{union}}$ using a fixed threshold as $\hat{\mathcal{D}}=\left\{\tilde{\mathbf r}_n: q_{\mathrm{union}}(\tilde{\mathbf r}_n)\geq0.5\right\}$.
	The IoU is defined as 
	\begin{equation} 
		\mathrm{IoU}=\frac{|\hat{\mathcal{D}}\cap\mathcal{D}|}{|\hat{\mathcal{D}}\cup\mathcal{D}|}. 
		\label{eq:IoU} 
	\end{equation} 
	A larger IoU indicates more accurate support recovery.  
	
	\subsection{Evolution of the Reconstruction During Inversion}
	\label{subsec:AnaofModel}
	To illustrate the inversion process, Figure~\ref{fig:qunion} shows the reconstructed relative permittivity distribution at different iterations for the multi-material Austria target. The corresponding IoU and $\mathrm{RE}_{\chi}$ values are listed below each iteration.
	
	At early iterations, the reconstruction still exhibits noticeable geometric errors and inaccurate permittivity values. As the inversion proceeds, the object boundaries become increasingly clear, the annular structure and the two smaller objects are gradually recovered, and the material values approach their ground truth values. The quantitative metrics show the same trend, with the IoU increasing from 0.6808 at $\nu=100$ to 0.9241 at $\nu=800$, while $\mathrm{RE}_{\chi}$ decreases from 0.6778 to 0.2851. These results demonstrate that the proposed method progressively improves both the recovered geometry and the complex relative permittivity distribution during the inversion.
	
	\begin{figure*}
		\centering
		\includegraphics[width=\linewidth]{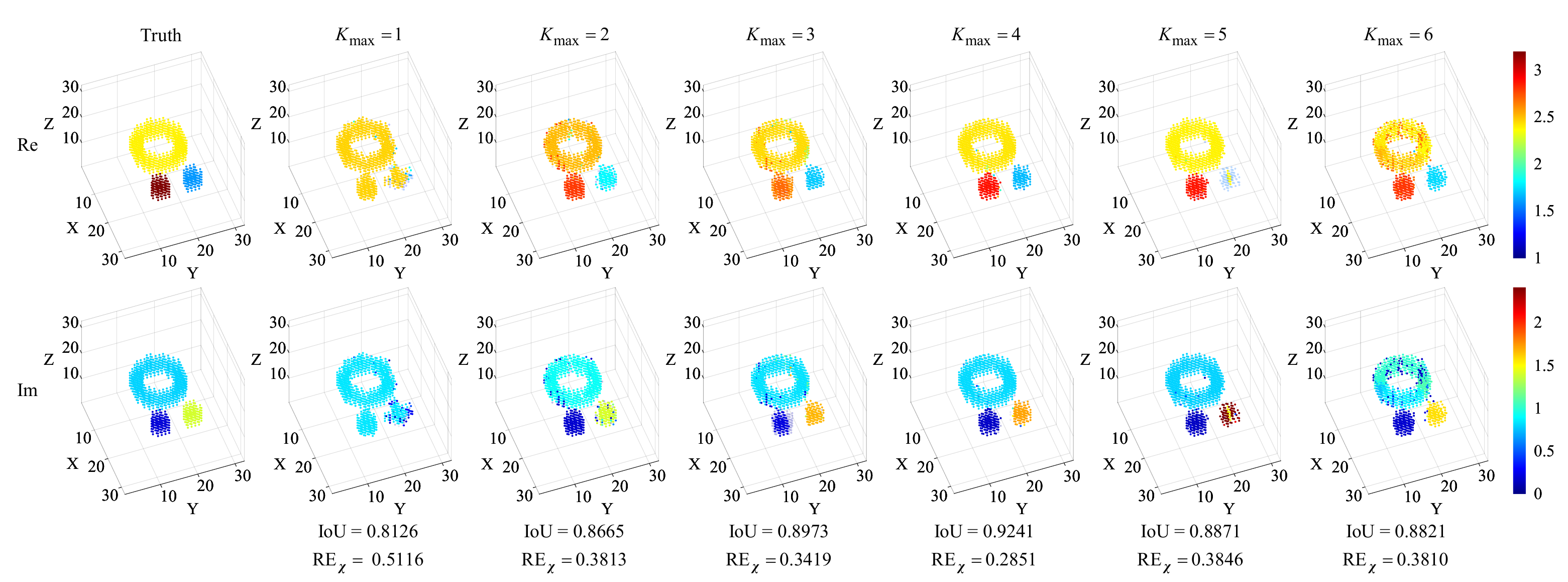}
		\caption{Sensitivity of the reconstruction to the number of level-set functions $K_{\max}$. The corresponding IoU and $\mathrm{RE}_{\boldsymbol{\chi}}$ values are given below.}
		\label{fig:KmaxSens}
	\end{figure*}
	
	\subsection{Sensitivity to the Number of Level-Set Functions}
	\label{subsec:SenstoCI}
	The parameter $K_{\max}$ determines the number of level-set functions, with each function associated with a trainable complex relative permittivity. Since a single level-set function can represent multiple separated regions, $K_{\max}$ does not need to match the number of objects to be reconstructed.
	
	To investigate the influence of $K_{\max}$, it is varied from 1 to 6 while all other reconstruction settings remain unchanged. The quantitative results and representative reconstructions are shown in Figure~\ref{fig:KmaxSens}.
	
	When $K_{\max}=1$, the reconstruction captures the overall target configuration but shows noticeable geometric distortion and inaccurate material values, indicating insufficient representation capacity. As $K_{\max}$ increases from 1 to 4, the object boundaries become better defined and the recovered material distribution becomes more consistent with the ground truth. The reconstruction is most accurate at $K_{\max}=4$, where the annular structure and the two smaller objects are clearly recovered, and the reconstructed material values are more accurate. This is also reflected in the quantitative results, with the highest IoU of $0.9241$ and the lowest $\mathrm{RE}_{\boldsymbol{\chi}}$ of $0.2851$. Increasing $K_{\max}$ to $5$ and $6$ does not improve the reconstruction. Instead, the IoU decreases while $\mathrm{RE}_{\boldsymbol{\chi}}$ increases. Although a larger $K_{\max}$ provides greater representation capacity, it also introduces additional level-set components and material parameters, increasing the complexity of the optimization. These results suggest that $K_{\max}$ should be sufficiently large to represent the target structure, but increasing it further does not necessarily improve reconstruction accuracy.
	
	\begin{figure*}
		\centering
		\includegraphics[width = \linewidth]{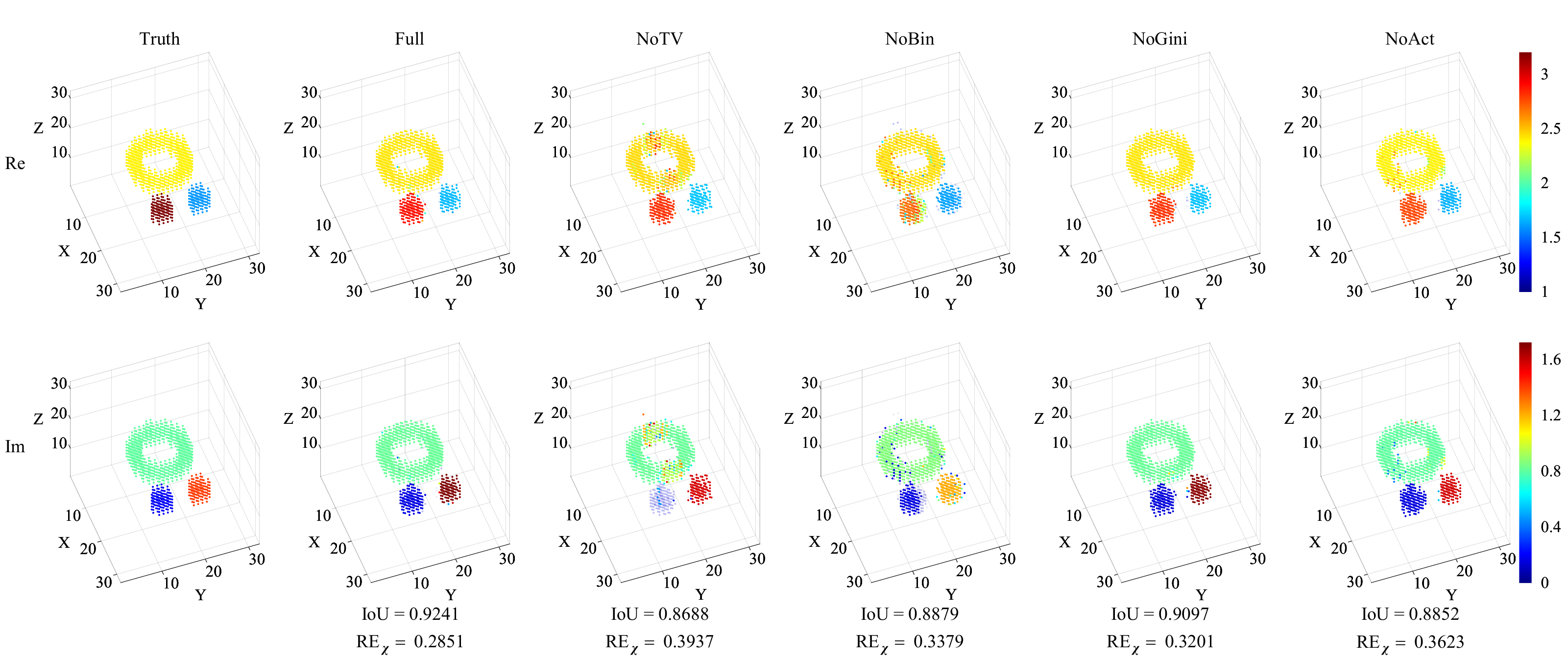}
		\caption{Ablation study of the regularization terms for the multi-material Austria target. Full denotes the complete SMLSI model, while NoTV, NoBin, NoGini, and NoAct denote the variants without $\mathcal R_{\mathrm{TV}}$, $\mathcal R_{\mathrm{Bin}}$, $\mathcal R_{\mathrm{Gini}}$ and $\mathcal R_{\mathrm{Act}}$, respectively. The corresponding IoU and $\mathrm{RE}_{\chi}$ values are given below.}
		\label{fig:Ablation}
	\end{figure*}
	
	\subsection{Evaluation of the Structured Representation}
	\label{subsec:AblationSR}
	To evaluate the benefit of replacing direct contrast parameterization with the proposed structured SMLSI formulation, a PDNN baseline, denoted as “PDNN (Direct $\chi$)”, is constructed using the same Fourier-feature encoding and MLP hidden-layer configuration as SMLSI, together with the same optimizer and maximum number of iterations. Unlike SMLSI, this baseline directly outputs the real and imaginary parts of the complex contrast at each spatial coordinate, following the physics-driven neural inversion formulation in~\cite{Du2025PDNN,Du2026IPDNN}.
	
	Two additional ablation variants are considered to evaluate the roles of the learnable gating mechanism and the soft-union operation. In the “Fixed gate” variant, the learnable gating mechanism is removed by fixing $\sigma(\eta_k)=1$ for all level-set functions throughout the inversion, while all other settings are kept unchanged. In the “Additive support” variant, the gates remain learnable, while the soft-union operation in~\eqref{eq:q_union} is replaced by direct summation of the weighted supports followed by clipping to $[0,1]$. The ``Full SMLSI'' corresponds to the complete proposed method.
	
	\begin{table}[!ht]
		\centering
		\caption{Comparison and ablation of the structured representation}
		\label{tab:AblationSR}
		\resizebox{.6\columnwidth}{!}{
			\begin{tabular}{l|c|c|c|c}
				\toprule
				\textbf{Metric} & \textbf{PDNN (Direct $\boldsymbol{\chi}$)} & \textbf{Fixed gate} & \textbf{Additive support} & \textbf{Full SMLSI} \\
				\midrule
				$\mathrm{RE}_{\boldsymbol{\chi}}$ & 0.5800 & 0.3823 & 0.3584 & 0.2851 \\
				IoU & -- & 0.8746 & 0.8866 & 0.9241 \\
				\bottomrule
			\end{tabular}
		}
	\end{table}
	
	The quantitative results are summarized in Table~\ref{tab:AblationSR}. SMLSI reduces $\mathrm{RE}_{\boldsymbol{\chi}}$ from $0.5800$ for PDNN (Direct $\boldsymbol{\chi}$) to $0.2851$, corresponding to a $50.8\%$ reduction in reconstruction error. Fixing all gates to $1$ increases $\mathrm{RE}_{\boldsymbol{\chi}}$ to $0.3823$ and reduces the IoU to $0.8746$, indicating the benefit of incorporating the learnable gating mechanism into the structured representation. Replacing the soft-union operation with direct summation followed by clipping to $[0,1]$ also degrades the reconstruction, yielding an $\mathrm{RE}_{\boldsymbol{\chi}}$ of $0.3584$ and an IoU of $0.8866$. This result indicates that the proposed soft-union operation provides a more effective way of combining the weighted soft supports than direct summation. In comparison, Full SMLSI achieves the lowest reconstruction error of $0.2851$ and the highest IoU of $0.9241$. These results demonstrate the effectiveness of the proposed structured multi-level-set representation and the contributions of the learnable gating mechanism and soft-union operation.
	
	\subsection{Ablation Study on the Loss Function}
	\label{subsec:Ablation}
	
	To evaluate the contribution of each regularization term, an ablation study is performed by removing one term at a time while keeping all other settings unchanged. The compared variants are NoTV, NoBin, NoGini, NoAct, and the full model.
	
	Figure~\ref{fig:Ablation} shows the reconstruction results for the multi-material Austria target. Full SMLSI achieves the best performance, with an IoU of $0.9241$ and an $\mathrm{RE}_{\boldsymbol{\chi}}$ of $0.2851$. Removing the TV term results in the largest degradation, reducing the IoU to $0.8688$ and increasing $\mathrm{RE}_{\boldsymbol{\chi}}$ to $0.3937$, with greater nonuniformity in the reconstructed material values within individual material regions. Without the binarization term, the reconstruction shows increased errors in both the object support and material values, with the IoU decreasing to $0.8879$ and $\mathrm{RE}_{\boldsymbol{\chi}}$ increasing to $0.3379$. Removing the Gini term also degrades the reconstruction, yielding an IoU of $0.9097$ and an $\mathrm{RE}_{\boldsymbol{\chi}}$ of $0.3201$. Removing the activation term also degrades the reconstruction, with an IoU of $0.8852$ and an $\mathrm{RE}_{\boldsymbol{\chi}}$ of $0.3623$. Overall, all four ablation variants perform worse than Full SMLSI, indicating that each regularization term contributes to the reconstruction performance for this representative case.
	
	\subsection{Evaluation of Adaptive Regularization Weighting}
	\label{subsec:EvaluARW}
	To evaluate the effectiveness of the adaptive regularization weighting strategy introduced in Section~\ref{subsec:OS}, we compare it with several fixed-weight settings obtained by scaling the initial regularization weights by different factors. For the fixed-weight variants, the initial weights obtained from \eqref{eq:lambda_init} are uniformly scaled as $\lambda_i=s\lambda_i^{(0)}$, where $s\in\{0.25,0.5,1,2,4\}$, and are then kept fixed throughout the subsequent inversion. The adaptive strategy starts from $\lambda_i^{(0)}$ at $\nu=\nu_w$ and updates the weights during the subsequent inversion.
	
	\begin{table}[!ht]
		\centering
		\caption{Quantitative comparison between fixed and adaptive regularization weighting strategies}
		\label{tab:adaptive_weight}
		\resizebox{.6\columnwidth}{!}{
			\begin{tabular}{l|c|c|c|c|c|c}
				\toprule
				\textbf{Metric}
				& $\boldsymbol{s=0.25}$
				& $\boldsymbol{s=0.5}$
				& $\boldsymbol{s=1}$
				& $\boldsymbol{s=2}$
				& $\boldsymbol{s=4}$
				& \textbf{Adaptive} \\
				\midrule
				$\mathrm{RE}_{\boldsymbol{\chi}}$
				& 0.3031 & 0.3285 & 0.2991
				& 0.3247 & 0.3489 & 0.2851 \\
				IoU
				& 0.9122 & 0.9010 & 0.9233
				& 0.9089 & 0.8992 & 0.9241 \\
				\bottomrule
			\end{tabular}
		}
	\end{table}
	
	\begin{figure}
		\centering
		\includegraphics[width = .58\linewidth]{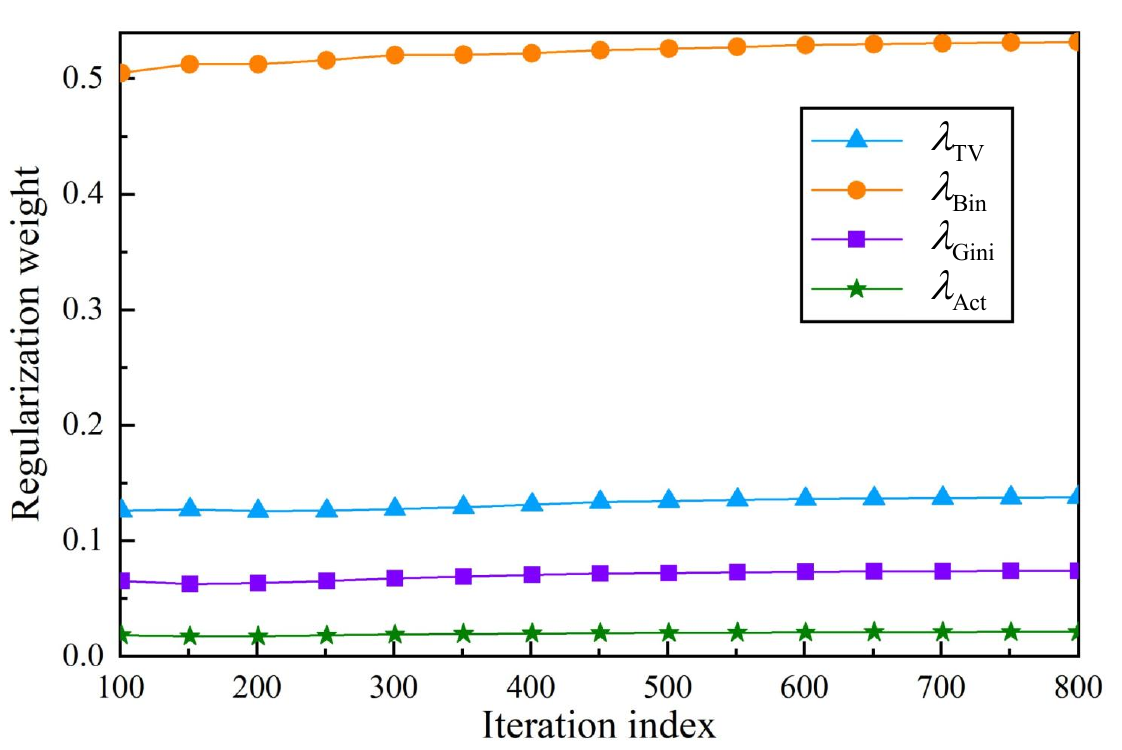}
		\caption{Evolution of the adaptively optimized regularization weights.}
		\label{fig:adaptive_weight}
	\end{figure}
	
	Table~\ref{tab:adaptive_weight} summarizes the quantitative results. For the fixed-weight variants, the reconstruction performance varies with the choice of $s$. Among them, $s=1$ achieves the best performance, with an $\mathrm{RE}_{\boldsymbol{\chi}}$ of 0.2991 and an IoU of 0.9233, whereas the other fixed settings lead to higher reconstruction errors and lower IoU values. These results show that the reconstruction performance varies noticeably with the choice of $s$ under fixed regularization weights. In comparison, the proposed adaptive weighting strategy achieves an $\mathrm{RE}_{\boldsymbol{\chi}}$ of 0.2851 and an IoU of 0.9241, outperforming all fixed-weight variants considered here. Figure~\ref{fig:adaptive_weight} shows the evolution of the four regularization weights during inversion. After initialization, the four weights evolve smoothly and gradually tend toward stable values. These results demonstrate that the adaptive weighting strategy achieves better reconstruction performance than the tested fixed-weight settings while reducing the dependence on manually selected regularization weights.
	
	\subsection{Ablation Study on the Optimization Strategy}
	\label{subsec:staged}
	
	\begin{figure}
		\centering
		\includegraphics[width = .6\linewidth]{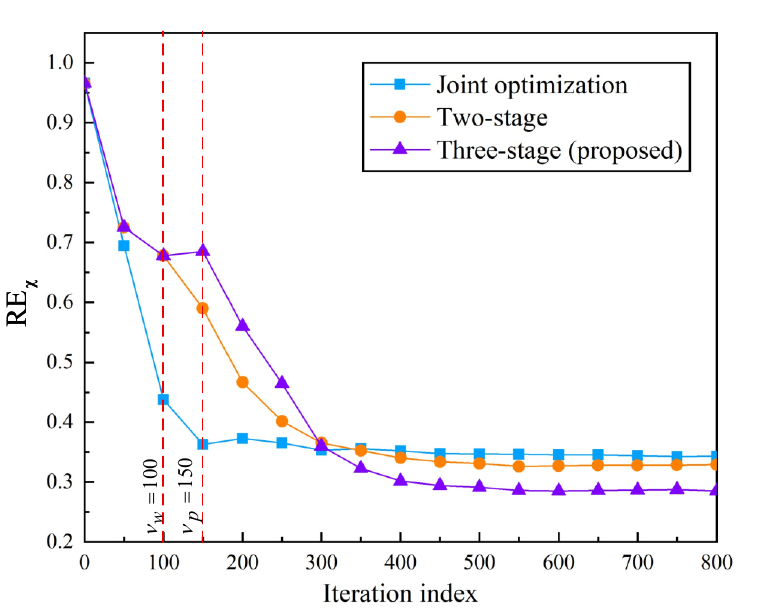}
		\caption{Evolution of $\mathrm{RE}_{\boldsymbol{\chi}}$ under different optimization strategies. The dashed lines mark the stage transitions at $\nu_w=100$ and $\nu_p=150$.}
		\label{fig:staged}
	\end{figure}
	
	To evaluate the proposed staged optimization strategy, three optimization schedules are compared while keeping all other reconstruction settings unchanged. In the ``Joint optimization'' strategy, all parameters are optimized with regularization from the beginning. In the ``Two-stage'' strategy, only the level-set and gating parameters are optimized before $\nu_w$. Regularization is introduced and relative permittivity optimization begins at $\nu_w$. In the proposed ``Three-stage'' strategy, regularization is introduced at $\nu_w$, while the relative permittivities are optimized from $\nu_p$ onward.
	
	Figure~\ref{fig:staged} compares the evolution of $\mathrm{RE}_{\boldsymbol{\chi}}$ for the three strategies. Joint optimization reduces the reconstruction error rapidly during the early iterations but converges to a higher final error. The two-stage strategy achieves a lower final error by keeping the relative permittivity parameters fixed and the regularization terms inactive until $\nu_w$. For the proposed three-stage strategy, $\mathrm{RE}_{\boldsymbol{\chi}}$ decreases substantially after the relative permittivity parameters are introduced into the optimization at $\nu_p$, eventually reaching the lowest final $\mathrm{RE}_{\boldsymbol{\chi}}$ among the three strategies. These results show the effectiveness of the proposed staged optimization strategy.
	
	\section{Reconstruction Performance and Benchmark Evaluation}
	\label{sec:Performance}
	Unless otherwise specified, the synthetic experiments in this section follow the numerical and optimization settings described in Section~\ref{subsec:paraSets}.
	
	\subsection{Representative Reconstruction Results}
	\label{subsec:RRR}
	\begin{figure*}[!t]
		\centering
		\includegraphics[width = .84\linewidth]{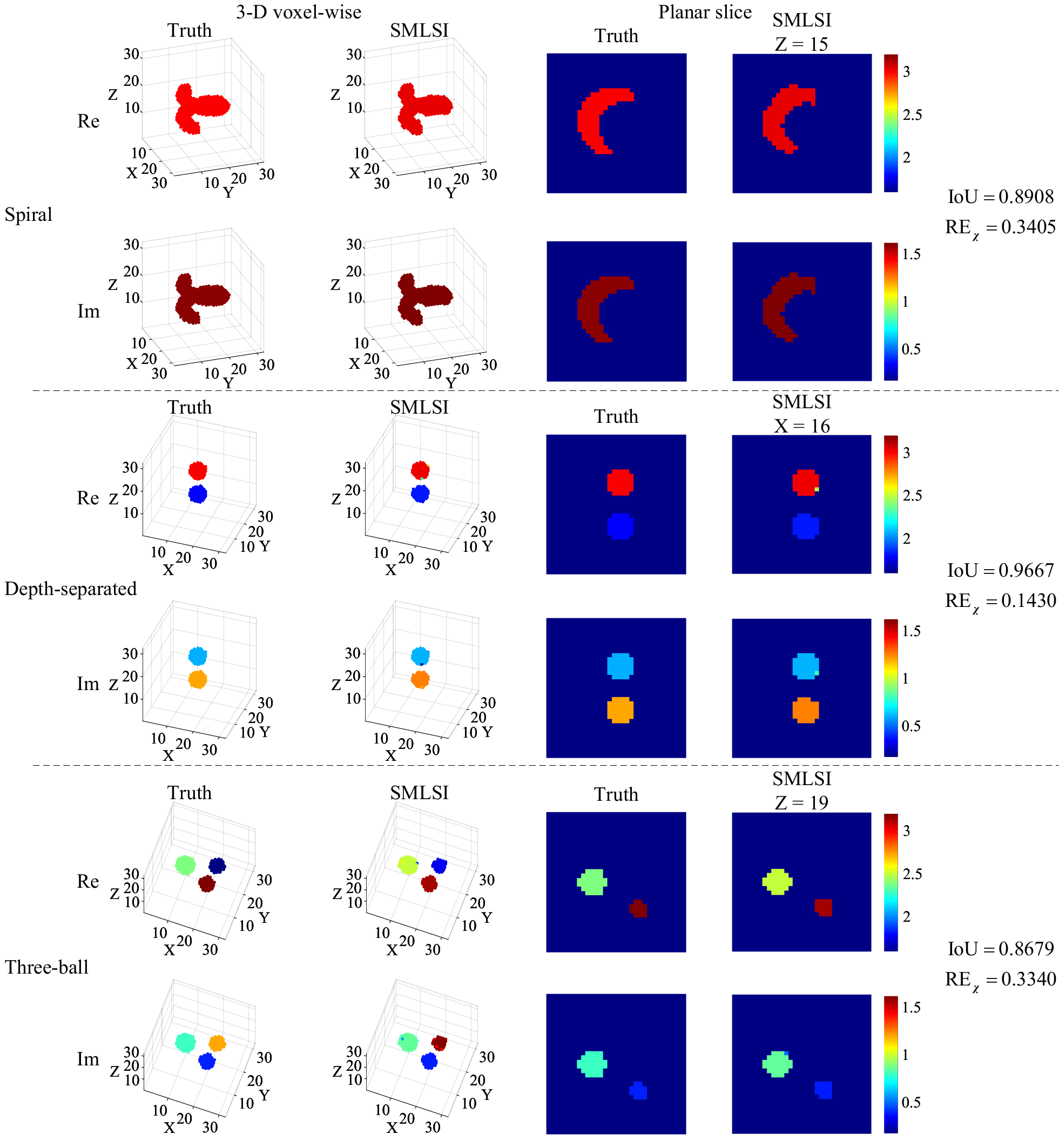}
		\caption{Representative 3-D reconstruction results of SMLSI for three target configurations. The corresponding IoU and $\mathrm{RE}_{\chi}$ values are given on the right.}
		\label{fig:RRR}
	\end{figure*}
	
	Three representative targets are considered to evaluate the reconstruction performance of SMLSI for different 3-D targets. The spiral target contains a spiral-shaped homogeneous scatterer and is used to examine the reconstruction of complex nonconvex boundaries. The depth-separated target consists of two scatterers with the same projection onto the $x$-$y$ plane but separated along the $z$-direction, testing the ability to resolve targets that overlap in projection but are separated in depth. The three-ball target contains three spatially separated scatterers with different complex permittivities and examines the ability to recover multiple targets and distinguish their material properties.
	
	Figure~\ref{fig:RRR} presents the reconstructed real and imaginary parts of the relative permittivity together with the corresponding quantitative results. For the spiral target, SMLSI recovers the main nonconvex geometry and preserves the irregular boundary in the selected cross section, achieving an IoU of $0.8908$ and an $\mathrm{RE}_{\boldsymbol{\chi}}$ of $0.3405$. For the depth-separated target, the two scatterers are clearly resolved along the $z$-direction, achieving an IoU of $0.9667$ and an $\mathrm{RE}_{\boldsymbol{\chi}}$ of $0.1430$. For the three-ball target, the three spatially separated scatterers are reconstructed at the correct locations, and their different complex relative permittivities are distinguished, yielding an IoU of $0.8679$ and an $\mathrm{RE}_{\boldsymbol{\chi}}$ of $0.3340$. Overall, these results demonstrate that SMLSI can reconstruct complex nonconvex geometries, resolve targets separated in depth, and recover multiple scatterers with different material properties. 
	
	\subsection{Reconstruction of Closely Spaced Scatterers}
	\label{subsec:Resolution}
	
	\begin{figure}[!t]
		\centering
		\includegraphics[width=.6\linewidth]{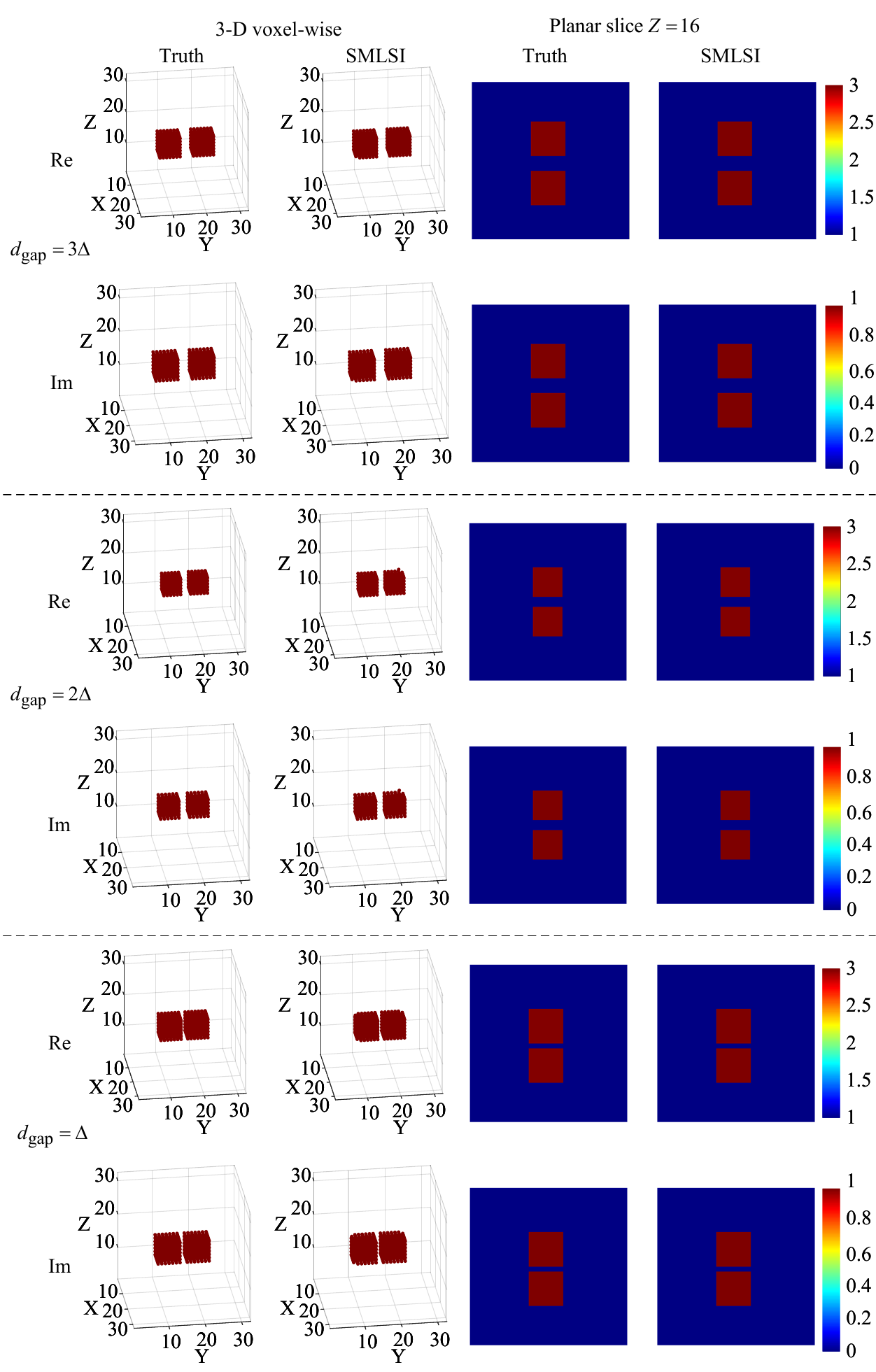}
		\caption{Reconstruction results of two closely spaced cubic scatterers with different separations $d_{\mathrm{gap}}=3\Delta$, $2\Delta$, and $\Delta$, where $d_{\mathrm{gap}}$ denotes the distance between the facing surfaces of the two cubes and $\Delta=\lambda/16$ is the voxel spacing corresponding to the discretization described in Section~\ref{subsec:paraSets}. }
		\label{fig:Resolution}
	\end{figure}
	
	To evaluate the ability of SMLSI to distinguish closely spaced scatterers, two cubic targets are considered. Let $\Delta=\lambda/16$ denote the voxel spacing corresponding to the discretization described in Section~\ref{subsec:paraSets}. Each cube has a side length of $7\Delta$, and the separation between their facing surfaces, denoted by $d_{\mathrm{gap}}$, is set to $\Delta$, $2\Delta$, and $3\Delta$.
	
	Figure~\ref{fig:Resolution} shows the reconstructed real and imaginary parts of the relative permittivity for the three separations. At $d_\text{gap}=3\Delta$ and $2\Delta$, the two scatterers are clearly separated, and their cubic geometries are accurately recovered. Even when the distance decreases to $d_\text{gap}=\Delta$, the two scatterers remain distinguishable.
	
	\begin{table}[!ht]
		\centering
		\caption{Quantitative reconstruction results for closely spaced cubic scatterers with different separations}
		\label{tab:Resolution}
		\resizebox{0.38\columnwidth}{!}{
			\begin{tabular}{l|c|c|c}
				\toprule
				\textbf{} & $d_\text{gap}=3 \Delta$ & $d_\text{gap}=2 \Delta$ & $d_\text{gap}=\Delta$ \\
				\midrule
				$\mathrm{RE}_{\boldsymbol{\chi}}$ & 0.0361 & 0.0766 & 0.0660 \\
				IoU & 0.9985 & 0.9931 & 0.9956 \\
				\bottomrule
			\end{tabular}
		}
	\end{table}
	
	The quantitative results are summarized in Table~\ref{tab:Resolution}. The IoU remains above $0.99$ for all three separations, indicating accurate recovery of the two target regions even at the smallest tested separation. The corresponding $\mathrm{RE}_{\boldsymbol{\chi}}$ values remain below $0.08$. These results are consistent with the visual reconstructions in Figure~\ref{fig:Resolution} and demonstrate that SMLSI maintains accurate geometric separation and material reconstruction for closely spaced scatterers.
	
	\subsection{Noise Robustness Analysis}
	\label{subsec:Noise}
	\begin{figure*}[!t]
		\centering
		\includegraphics[width=\linewidth]{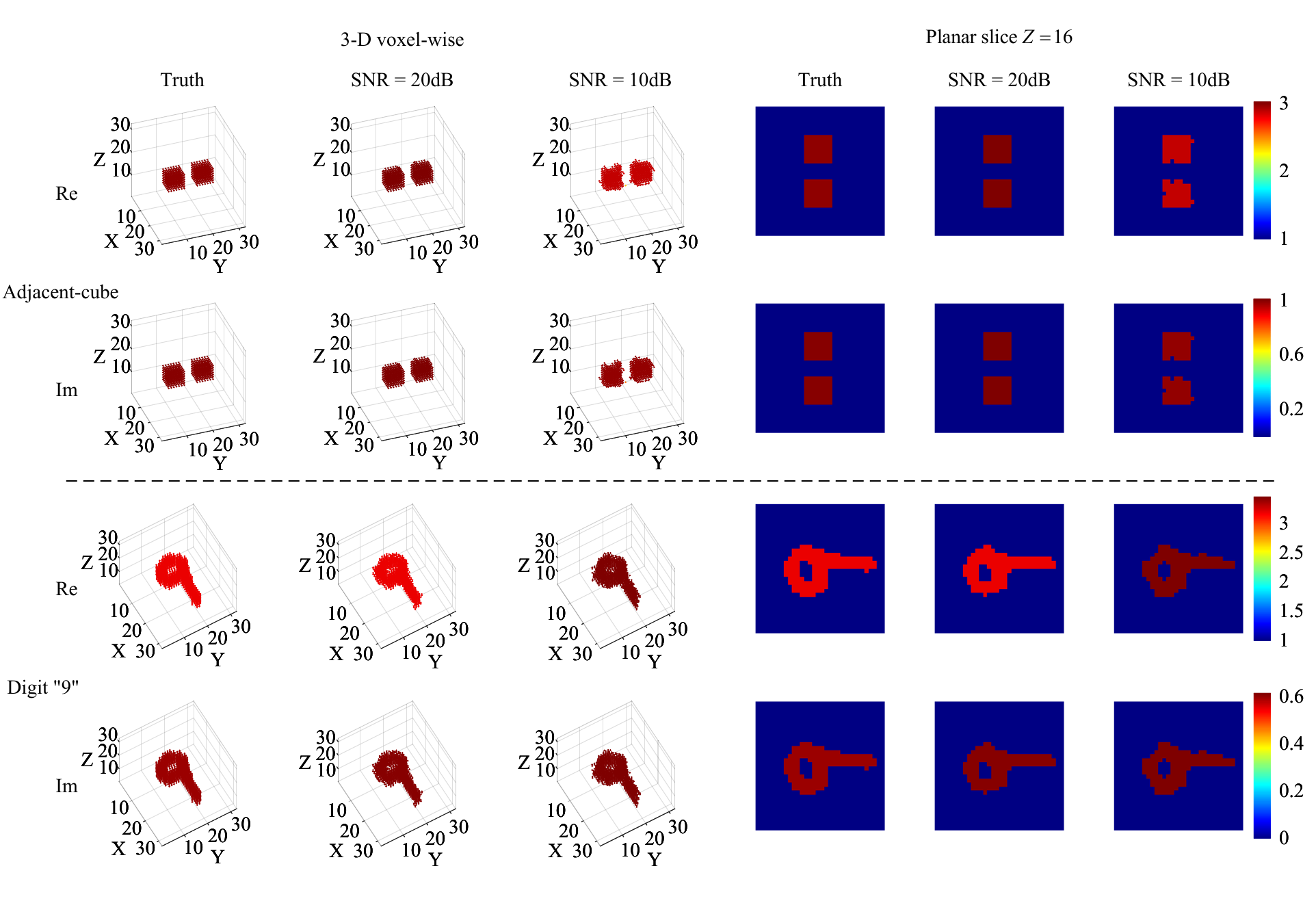}
		\caption{Reconstruction results of SMLSI under noisy measurements with SNRs of $20$~dB and $10$~dB. Two representative targets are considered.}
		\label{fig:Noise}
	\end{figure*}
	To evaluate the robustness of SMLSI against measurement noise, additive white Gaussian noise (AWGN) is introduced into the scattered field data at SNR levels of $20$~dB and $10$~dB. Two representative targets with different geometric configurations are considered, and the corresponding reconstruction results are shown in Figure~\ref{fig:Noise}.
	
	For the adjacent-cube target, the two scatterers are accurately recovered at $20$~dB, with clear boundaries and nearly uniform material values. When the SNR decreases to $10$~dB, increased boundary distortion and spatial nonuniformity can be observed, particularly in the planar slices, but the two target regions remain correctly localized and separated. For digit ``9'', the main object geometry is also preserved under both noise levels. The degradation becomes more evident at $10$~dB, where larger deviations appear in the reconstructed material values, although the principal structural features remain identifiable.
	\begin{table}[!ht]
		\centering
		\caption{Quantitative reconstruction results under different noise levels}
		\label{tab:Noise}
		\resizebox{0.38\columnwidth}{!}{
			\begin{tabular}{lcccc}
				\toprule
				& \multicolumn{2}{c}{\textbf{Adjacent-cube}}
				& \multicolumn{2}{c}{\textbf{Digit ``9''}} \\
				\cmidrule(lr){2-3}
				\cmidrule(lr){4-5}
				& \textbf{20 dB} & \textbf{10 dB}
				& \textbf{20 dB} & \textbf{10 dB} \\
				\midrule
				$\mathrm{RE}_{\boldsymbol{\chi}}$
				& 0.1814 & 0.4213 & 0.4397 & 0.5096 \\
				IoU
				& 0.9684 & 0.8348 & 0.8250 & 0.7915 \\
				\bottomrule
			\end{tabular}
		}
	\end{table}
	
	The quantitative results in Table~\ref{tab:Noise} are consistent with the visual observations. For the adjacent-cube target, decreasing the SNR from $20$~dB to $10$~dB increases $\mathrm{RE}_{\boldsymbol{\chi}}$ from $0.1814$ to $0.4213$, while the IoU decreases from $0.9684$ to $0.8348$. For digit ``9'', $\mathrm{RE}_{\boldsymbol{\chi}}$ increases from $0.4397$ to $0.5096$, while the IoU decreases from $0.8250$ to $0.7915$. These results indicate that the reconstruction accuracy degrades as the noise level increases, while the main structural features remain preserved at $10$~dB.
	
	\subsection{Comparison With Existing Inversion Methods}
	\label{subsec:compareMethods}
	
	\begin{figure*}[!t]
		\centering
		\includegraphics[width=\linewidth]{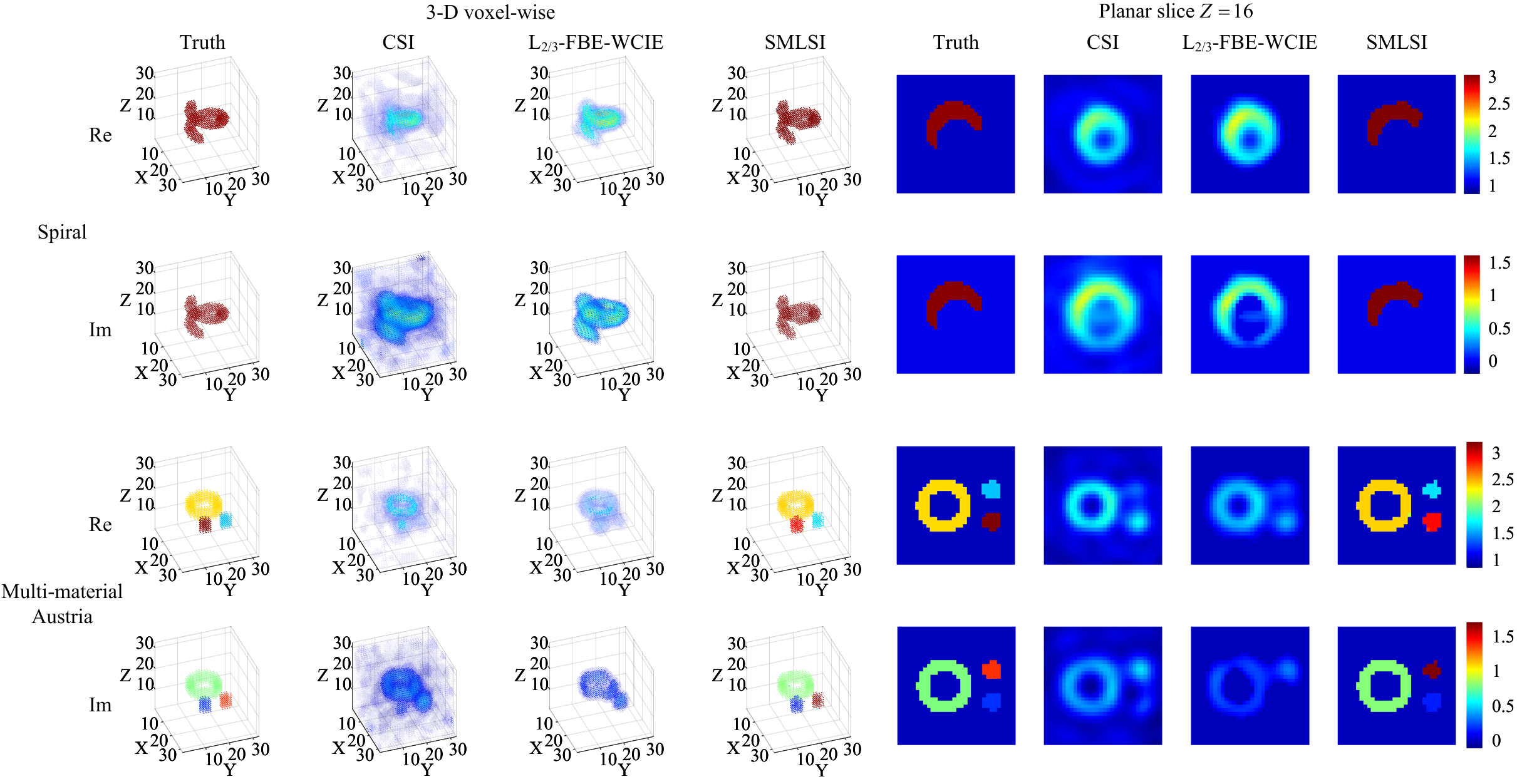}
		\caption{Comparison of the reconstruction results obtained by CSI, $L_{2/3}$-FBE-WCIE, and the proposed SMLSI for two representative 3-D cases.}
		\label{fig:compare}
	\end{figure*}
	
	To evaluate the reconstruction performance of SMLSI against existing inversion methods, the conventional contrast source inversion (CSI) method and the recently developed $L_{2/3}$-FBE-WCIE method~\cite{Li2026FBE3D} are selected as baselines. Two representative 3-D cases with different geometric and material configurations are considered. All methods use the same scattered field data and measurement configuration.
	
	Figure~\ref{fig:compare} compares the reconstructed real and imaginary parts of the relative permittivity. For the spiral target, CSI identifies the rough target region but produces noticeable background artifacts. Benefiting from the $L_{2/3}$-norm sparsity regularization, the $L_{2/3}$-FBE-WCIE method substantially suppresses background artifacts compared with CSI. However, the reconstructed boundaries and material interfaces are still not clearly defined. In contrast, SMLSI better preserves the overall target geometry and produces more clearly defined object boundaries. The corresponding planar slices further show accurate recovery of the boundary shape and permittivity values. For the multi-material Austria target, which contains multiple material components with different complex permittivities, the differences among the three methods become more evident. CSI recovers the approximate target locations, but the reconstructed material regions are strongly affected by background artifacts, making it difficult to distinguish the different permittivity values. The $L_{2/3}$-FBE-WCIE method substantially suppresses the background artifacts owing to the sparsity regularization. However, although the sparsity prior is effective for artifact suppression, it may also over-penalize weak but physically meaningful material components. In particular, one component in the imaginary part is strongly suppressed and is barely visible in the reconstruction, resulting in an incomplete recovery of the multi-material distribution. In addition, the reconstructed permittivity values remain noticeably biased from the ground truth. In contrast, SMLSI preserves all material components and their spatial distribution, while providing a clearer separation of the different permittivity values. The planar slices further show that both the material interfaces and the corresponding real and imaginary permittivity values are more accurately recovered by SMLSI.
	
	\begin{table}[!ht]
		\centering
		\caption{Quantitative comparison of different inversion methods for the two representative 3-D cases}
		\label{tab:compare}
		\resizebox{.52\columnwidth}{!}{
			\begin{tabular}{lcccc}
				\toprule
				& \multicolumn{2}{c}{\textbf{Spiral}}
				& \multicolumn{2}{c}{\textbf{Multi-material Austria}} \\
				\cmidrule(lr){2-3}\cmidrule(lr){4-5}
				\textbf{Method}
				& $\mathrm{RE}_{\boldsymbol{\chi}}$ & Runtime (s)
				& $\mathrm{RE}_{\boldsymbol{\chi}}$ & Runtime (s) \\
				\midrule
				CSI                 & 0.7799 & 883  & 0.7383 & 910  \\
				$L_{2/3}$-FBE-WCIE & 0.7263 & 1580 & 0.7770 & 1796 \\
				SMLSI              & 0.3405 & 238  & 0.2851 & 322  \\
				\bottomrule
			\end{tabular}
		}
	\end{table}
	
	The quantitative results in Table~\ref{tab:compare} further support the visual comparison. For the spiral target, SMLSI achieves an $\mathrm{RE}_{\boldsymbol{\chi}}$ of $0.3405$, compared with $0.7799$ for CSI and $0.7263$ for $L_{2/3}$-FBE-WCIE. For the multi-material Austria target, the corresponding error is reduced to $0.2851$, whereas CSI and $L_{2/3}$-FBE-WCIE yield errors of $0.7383$ and $0.7770$, respectively. In addition to the lower reconstruction errors, SMLSI requires only $238$~s and $322$~s for the two cases, substantially less than the two baseline methods. Averaged over the two cases, SMLSI achieves an $\mathrm{RE}_{\boldsymbol{\chi}}$ of $0.3128$, compared with $0.7591$ for CSI and $0.7517$ for $L_{2/3}$-FBE-WCIE. The corresponding average runtime is $280$~s, which is approximately $3.2\times$ faster than CSI and $6.0\times$ faster than $L_{2/3}$-FBE-WCIE. These results indicate that the proposed SMLSI method provides improved reconstruction accuracy while maintaining substantially lower runtime for the considered 3-D ISPs.
	
	\subsection{Experimental Validation}
	\label{subsec:ExpVal}
	\begin{figure}
		\centering
		\includegraphics[width = .38\linewidth]{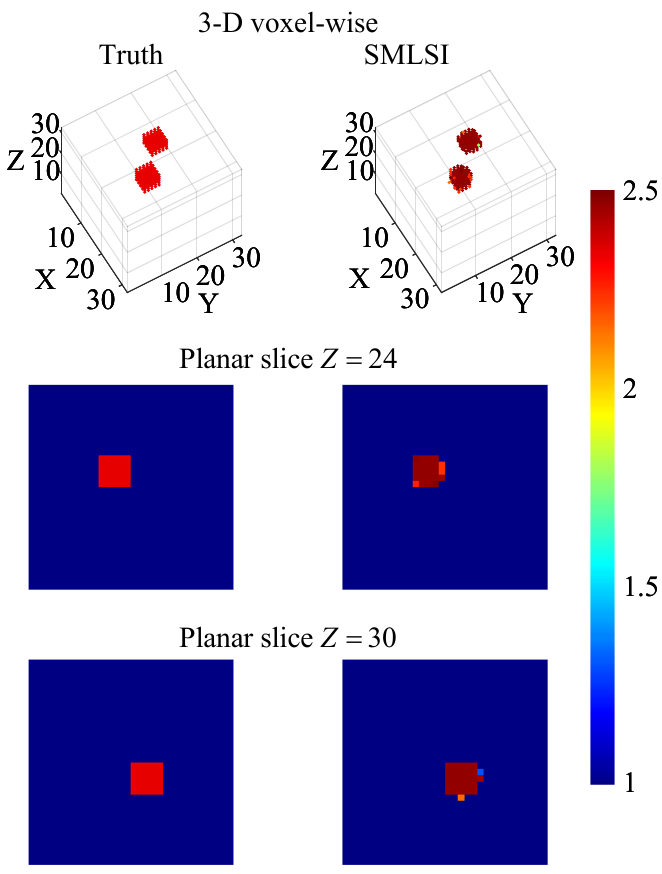}
		\caption{Imaging results of the proposed SMLSI using the experimental ``TwoCubes'' measurements~\cite{Geffrin2009}.}
		\label{fig:ExpTwoCubes}
	\end{figure}
	
	\begin{figure}
		\centering
		\includegraphics[width = .38\linewidth]{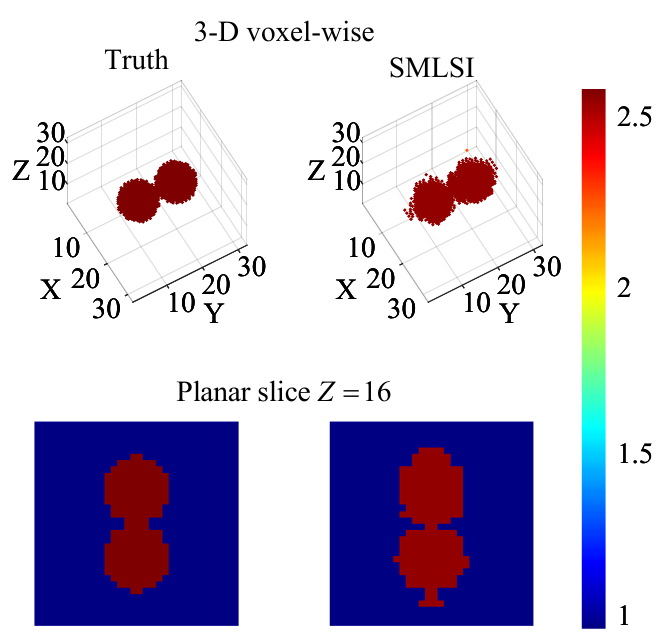}
		\caption{Imaging results of the proposed SMLSI using the experimental ``TwoSpheres'' measurements~\cite{Geffrin2009}.}
		\label{fig:ExpTwoSpheres}
	\end{figure}
	
	To further evaluate the applicability of SMLSI to measured data, experimental validation is performed using the open-access 3-D Fresnel database~\cite{Geffrin2009}. The database provides both co-polarized $\phi\phi$ and cross-polarized $\theta\phi$ measurements. In the following experiments, only the co-polarized $\phi\phi$ measurements, denoted as the PP channel, are used for inversion.
	
	The ``TwoCubes'' and ``TwoSpheres'' targets are considered at an operating frequency of $6$~GHz. The ``TwoCubes'' target consists of two dielectric cubes with a side length of $25$~mm and a relative permittivity of $2.3$. This configuration is used to examine the recovery of separated objects with sharp edges. The ``TwoSpheres'' target consists of two dielectric spheres with a diameter of $50$~mm and a relative permittivity of $2.6$, providing an evaluation scenario characterized by smooth curved boundaries and two adjacent objects.
	
	Figure~\ref{fig:ExpTwoCubes} presents the reconstruction results for the ``TwoCubes'' target. SMLSI successfully identifies two separated target regions at the expected spatial locations. The 3-D reconstruction preserves the overall cubic structures, while the planar slices at $z=24$ and $z=30$ further confirm the localization of the two individual scatterers. Although slight boundary distortions from the ideal cubic shape are visible, the two scatterers remain clearly separated, with their spatial locations and sizes well recovered. Figure~\ref{fig:ExpTwoSpheres} shows the corresponding results for the ``TwoSpheres'' target. The two adjacent spherical regions are successfully reconstructed, and their relative positions and overall spatial arrangement agree with the reference geometry. The representative $z=16$ slice further shows that the adjacency of the two objects is preserved. Some surface roughness and local boundary deformation can be observed in the reconstructed result, which may reflect the combined effects of measurement errors and modeling discrepancies. Nevertheless, the main support and overall geometry of the two spheres remain identifiable.
	
	\begin{table}[!ht]
		\centering
		\caption{Quantitative reconstruction results for the experimental measurements}
		\label{tab:exp}
		\resizebox{0.3\columnwidth}{!}{
			\begin{tabular}{l|c|c}
				\toprule
				\textbf{} & TwoCubes & TwoSpheres \\
				\midrule
				$\mathrm{RE}_{\boldsymbol{\chi}}$ &0.4255  &0.4939  \\
				IoU &0.8264  &0.7886  \\
				\bottomrule
			\end{tabular}
		}
	\end{table}
	
	The quantitative results are summarized in Table~\ref{tab:exp}. For the ``TwoCubes'' target, SMLSI achieves an $\mathrm{RE}_{\boldsymbol{\chi}}$ of $0.4255$ and an IoU of $0.8264$. For the ``TwoSpheres'' target, the corresponding values are $0.4939$ and $0.7886$, respectively. The relatively high IoU values are consistent with the visual results, indicating that the main target regions and their spatial locations are successfully recovered from the experimental measurements.
	
	Overall, the experimental results show that SMLSI can be applied to measured 3-D scattering data rather than only to synthetic data generated by the forward model. Using only the PP measurement channel, SMLSI successfully reconstructs both the separated cubic targets and the adjacent spherical targets. Although some deviations remain in the reconstructed material values and local boundaries, the main target supports and spatial configurations are consistently recovered. These results provide experimental evidence for the applicability of the proposed structured multi-level-set inversion (SMLSI) framework to 3-D inverse scattering.
	
	\section{Conclusion}
	\label{sec:Conclu}
	This paper proposed a structured multi-level-set inversion (SMLSI) method for 3-D electromagnetic inverse scattering. By representing the unknown contrast through object support and material distribution rather than directly parameterizing voxel-wise contrast values, SMLSI incorporates the piecewise-homogeneous structure of dielectric targets into the inversion. The inversion is performed directly from scattered field measurements within a physics-driven neural network framework. Structure-aware regularization and adaptive regularization weighting are incorporated into the optimization.
	
	The numerical results demonstrate that the structured representation improves reconstruction accuracy over direct contrast parameterization and enables the recovery of complex geometries, multiple material regions, and closely spaced scatterers. SMLSI also retains the main target structures under noisy measurements. In the benchmark comparisons, it achieves an average $\mathrm{RE}{\boldsymbol{\chi}}$ of $0.3128$, compared with $0.7591$ for CSI and $0.7517$ for $L_{2/3}$-FBE-WCIE, while requiring shorter runtimes in the present implementation. The results obtained from the measured 3-D Fresnel data further demonstrate the applicability of the proposed method to experimental inverse scattering. Future work will investigate SMLSI under more realistic 3-D electromagnetic sensing configurations, with particular attention to robustness against model mismatch and measurement uncertainties, adaptive selection of the representation capacity, and validation on more complex experimental data.
	
	\bibliographystyle{unsrt}
	\bibliography{References}
\end{document}